\documentclass[conference]{IEEEtran}
\IEEEoverridecommandlockouts

\usepackage{cite}                  
\usepackage{amsmath,amssymb,amsfonts} 
\usepackage{algorithmic}            
\usepackage{graphicx}               
\usepackage{textcomp}               
\usepackage{xcolor}                 
\usepackage{caption}                
\usepackage{newtxtext,newtxmath}    
\usepackage{multirow}               
\usepackage{arydshln}               
\usepackage{comment}                
\def\BibTeX{{\rm B\kern-.05em{\sc i\kern-.025em b}\kern-.08em
    T\kern-.1667em\lower.7ex\hbox{E}\kern-.125emX}}

\newcommand{\bl}[1]{\textbf{#1}}

\newcommand{\mc}[1]{\mathcal{#1}}

\DeclareSymbolFont{cmletters}{OML}{cmm}{m}{it} 
\DeclareMathSymbol{\cmg}{\mathord}{cmletters}{`g}
\begin{document}

\def\ie{{\em i.e.}}
\def\eg{{\em e.g.}}
\def\etal{{\em et al.}}

\graphicspath{{./fig/}}

\title{Distilling Generative-Discriminative Representations for Very Low-Resolution Face Recognition}

\author{\IEEEauthorblockN{Junzheng Zhang\IEEEauthorrefmark{1}\IEEEauthorrefmark{2}, 
Weijia Guo\IEEEauthorrefmark{1}\IEEEauthorrefmark{2}, 
Bochao Liu\IEEEauthorrefmark{1}\IEEEauthorrefmark{2}, 
Ruixin Shi\IEEEauthorrefmark{1}\IEEEauthorrefmark{2}, 
Yong Li\IEEEauthorrefmark{1}\IEEEauthorrefmark{2}, 
and Shiming Ge\IEEEauthorrefmark{1}\IEEEauthorrefmark{2}}
\IEEEauthorblockA{\IEEEauthorrefmark{1}Institute of Information Engineering, Chinese Academy of Sciences, Beijing 100085, China}
\IEEEauthorblockA{\IEEEauthorrefmark{2}School of Cyber Security, University of Chinese Academy of Sciences, Beijing 100090, China\\
Email: \{zhangjunzheng, guoweijia, liubochao, shiruixin, liyong, geshiming\}@iie.ac.cn}
\thanks{Thanks to the grant from the Pioneer R\&D Program of Zhejiang Province (2024C01024) for funding. Shiming Ge is the corresponding author.}}

\maketitle

\begin{abstract}
Very low-resolution face recognition is challenging due to the serious loss of informative facial details in resolution degradation. Recent approaches based on knowledge distillation provide an effective solution by distilling knowledge from a well-trained teacher for high-resolution face recognition and transferring it to a student for low-resolution face recognition. In general, the existing approaches usually take a discriminative model as teacher, where the teacher knowledge is trained in an abstract manner and provides poor transfer efficiency to compensate for the missing knowledge in low-resolution faces. To make more complete knowledge transfer, we propose a generative-discriminative representation distillation approach that combines generative representation with cross-resolution aligned knowledge distillation. This approach facilitates very low-resolution face recognition by jointly distilling generative and discriminative models via two distillation modules. Firstly, the generative representation distillation takes the encoder of a diffusion model pretrained for face super-resolution as the generative teacher to supervise the learning of the student backbone via feature regression, and then freezes the student backbone. After that, the discriminative representation distillation further considers a pretrained face recognizer as the discriminative teacher to supervise the learning of the student head via cross-resolution relational contrastive distillation. In this way, the general backbone representation can be transformed into discriminative head representation, leading to a robust and discriminative student model for very low-resolution face recognition. Our approach improves the recovery of the missing details in very low-resolution faces and achieves better knowledge transfer. Extensive experiments on face datasets demonstrate that our approach enhances the recognition accuracy of very low-resolution faces, showcasing its effectiveness and adaptability.
\end{abstract}

\begin{IEEEkeywords}
Very low-resolution face recognition, generative representation, knowledge distillation
\end{IEEEkeywords}

\section{Introduction}
Low-resolution face recognition is critical in many practical applications like remote video surveillance in the wild~\cite{gong2019iccvw, chai2023cvpr}. During the resolution degradation of normal high-resolution faces into a very low resolution (e.g., $16 \times 16$), a lot of informative details are often missing, which challenges the traditional state-of-the-art face recognizers~\cite{schroff2015facenet,wang2018cvpr,deng2019arcface,kim2020groupface} whose recognition accuracy is usually greatly reduced~\cite{ge2020efficient}. Thus, the key challenge in low-resolution face recognition is to transfer or generate useful knowledge to compensate for the lost information. To address that, the existing approaches can be grouped into two main categories according to the transfer or generation idea.

The transfer-based approaches aim to transfer the knowledge from high-resolution images to low-resolution recognition models for learning. Early approach~\cite{biswas2011multidimensional} proposed to directly learn the feature embedding from low-resolution faces. Recently, approaches based on knowledge distillation \cite{ge2019low, zhu2019icassp, yan2019iccvw, ge2019mm, ge2020efficient, ge2020aaai, zhang2024tcsvt} have achieved success in the field of low-resolution face recognition by transferring knowledge from teacher to student. However,  their transfer efficiency is still poor since their teacher knowledge is extracted in an abstract manner, while low-resolution face recognition needs more complete knowledge for transfer. By contrast, the generation-based approaches aim to complete the missing pixel-level or latent-level information. Early approach~\cite{zha2019tcn} enhanced the facial features of low-resolution faces through the super-resolution process, making them easier to match. Recently, generative models~\cite{rombach2022cvpr,yang2023nips,esser2024icml} learned from large-scale datasets have demonstrated efficient image generation and super-resolution capabilities. Some recent works~\cite{yang2023iccv, ge2024icml} integrate generative models into representation learning, and enhance low-quality visual recognition tasks.

In this paper, we propose a generative-discriminative representation distillation approach to facilitate very low-resolution face recognition in a progressive training manner via two distillation modules. Firstly, the generative representation distillation stabilizes the student backbone by distilling the encoder of a face super-resolution diffusion model and performing feature regression on the generative features, thus completing the knowledge of very low-resolution faces at the latent level. Secondly, the discriminative representation distillation finetunes the student head by distilling a pretrained high-resolution face recognizer with cross-resolution relational contrastive distillation. This process enables the generative features to continuously approximate the discriminative features, allowing the student model to learn the teacher model's knowledge more accurately. Our approach helps the student recover the missing latent details, achieves precise knowledge transfer, and enhances recognition accuracy. The main contributions include:~1) we propose a generative-discriminative representation distillation approach to promote very low-resolution face recognition; 2) we propose a progressive and module-wise approach for efficient student training; 3) we conduct extensive experiments on four popular benchmarks to validate the effectiveness of our approach. 
\begin{small}
\begin{figure*}[htbp]
\centering
\includegraphics[width=1\linewidth]{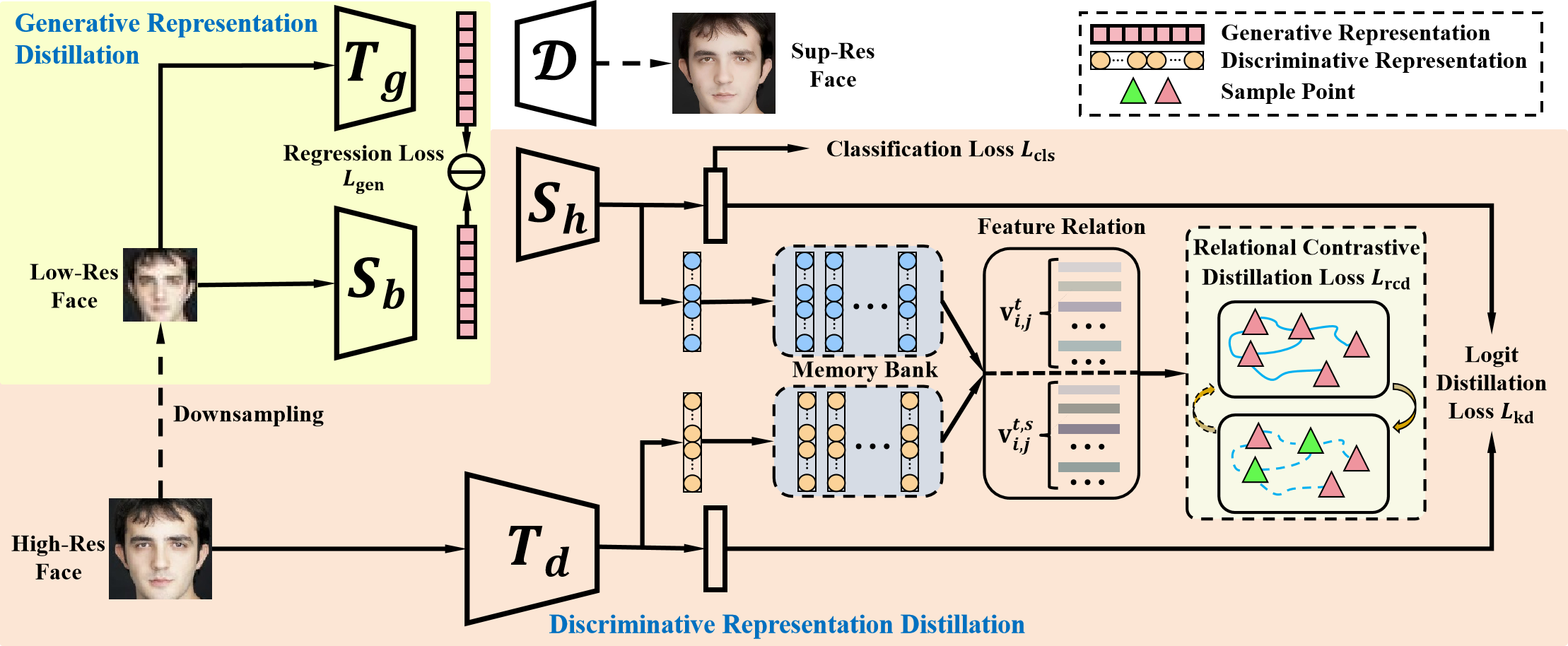} 
\caption{Our generative-discriminative representation distillation (GDRD) progressively trains the student \(S=\{S_b,S_h\}\) via two distillation modules. The generative representation distillation trains and freezes the student backbone \(S_b\) by distilling the encoder of a pretrained generative teacher \(T_{\cmg}\) via feature regression, and the discriminative representation distillation further trains the student head \(S_h\) by distilling a pretrained discriminative teacher \(T_d\) via relational contrastive distillation.}
\label{fig:framework}
\end{figure*}
\end{small}

\section{Approach}
Our objective is learning a low-resolution student $S$ on the training set $\mc{D}=\{\hat{\bl{x}}_i, \bl{x}_i, y_i\}_{i=1}^{|\mc{D}|}$ by distilling 1) general knowledge from a generative teacher $T_{\cmg}(\hat{\bl{x}}_i;\bl{w}_{\cmg})$ pretrained for super-resolving low-resolution face $\hat{\bl{x}}_i$ and 2) specific knowledge from a discriminative teacher $T_d(\bl{x}_i;\bl{w}_d)$ pretrained for recognizing high-resolution face $\bl{x}_i$. Here, \(y_i \in \{1, 2, \ldots, c\}\) represents face label in total $c$ classes, and \(\bl{w}_{\cmg}\) and \(\bl{w}_d\) are the weights of $T_{\cmg}$ and $T_d$ respectively. We separate the student \(S=\{S_b, S_h\}\) into a backbone \(S_b(\hat{\bl{x}}_i;\bl{w}_b)\) with parameters $\bl{w}_b$ to extract the intermediate feature \(\bl{f}_i\) and a head \(S_h(\bl{f}_i;\bl{w}_h)\) with parameters $\bl{w}_h$ to classify the feature, and then progressively perform two distillation modules, as shown in Fig. \ref{fig:framework}.
\subsection{Generative Representation Distillation}
It aims to enable the student to mimic the general representation ability of the generative teacher. We use the encoder of a pretrained diffusion model PGDiff~\cite{yang2023nips} as the generative teacher \(T_{\cmg}\). PGDiff simulates key properties of high-quality images, and is effective for super-resolution process, providing robust representational knowledge transfer due to its independence from the degradation process. During training, we perform feature regression by minimizing the generative loss \(L_{\text{gen}}\):
\begin{small}
\begin{equation}
  L_{\text{gen}}(\bl{w}_b;\mc{D}) = \sum_{t=1}^{|\mc{D}|} \left\| S_b(\hat{\bl{x}}_i;\bl{w}_b)-T_{\cmg}(\hat{\bl{x}}_i;\bl{w}_{\cmg}) \right\|^2.
\label{eq:genloss}
\end{equation}
\end{small}
\subsection{Discriminative Representation Distillation}
It aims to transform the general generative representations into discriminative ones for specific low-resolution recognition. Thus, we incorporate a well-trained state-of-the-art face recognizer ArcFace~\cite{deng2019arcface} as the dicriminative teacher and transfer its knowledge to learn the student head \(S_h\). To make the transfer effective, we leverage cross-resolution relational contrastive distillation that has been proved effective~\cite{zhang2024tcsvt}. 

Let \(\bl{x}\) and \(\bl{f}=S_b(\hat{\bl{x}};\bl{w}_b)\) respectively represent the inputs for \(T_d\) and \(S_h\), and their empirical data distribution is denoted as \(p(\bl{x},\bl{f})\). Then, we define the sampling procedure as \(p(\bl{x}, \bl{f})\sim(\bl{x}_i, \bl{x}_j, \bl{f}_i,\bl{f}_j) \), and set up two learnable feature relation modules (\(F^t\) and \(F^{t,s}\)) to represent the relationships of sample pairs as vectors
\(\bl{v}_{i,j}^t = F^t(T_d(\bl{x}_i;\bl{w}_d), T_d(\bl{x}_j;\bl{w}_d))\) in the teacher space and \(\bl{v}_{i,j}^{t,s} = F^{t,s}(T_d(\bl{x}_i;\bl{w}_d), S_h(\bl{f}_j; \bl{w}_b))\) in the cross-resolution space, respectively. We further set the feature transformations \(h_1\) and \(h_2\), and define 
the relational contrastive loss \(L_{\text{rcd}}\):
\begin{small}

\begin{equation}
\begin{aligned}
L_{\text{rcd}}& = -\sum_{\mc{P}(b=1)} \log \left(\frac{e^{h_1(\bl{v}_{i,j}^t)\cdot h_2(\bl{v}_{i,j}^{t,s})/\tau}}{e^{h_1(\bl{v}_{i,j}^t)\cdot h_2(\bl{v}_{i,j}^{t,s})/\tau} + n/m}\right) \\
&- \sum_{\mc{P}(b=0)} \log \left(1 - \frac{e^{h_1(\bl{v}_{i,j}^t)\cdot h_2(\bl{v}_{i,j}^{t,s})/\tau}}{e^{h_1(\bl{v}_{i,j}^t)\cdot h_2(\bl{v}_{i,j}^{t,s})/\tau} + n/m}\right),
\label{eq:rcd}
\end{aligned}
\end{equation}
\end{small}
where \(n\) and \(m\) are the number of negative sample pairs and total sample pairs during a training batch, respectively, and we set the same number of positive and negative sample pairs. We set the temperature parameter \(\tau=0.4\) to adjust the scale in our experiments. \(\mc{P}(b = 1)\) and \(\mc{P}(b = 0)\) denote the set of positive and negative sample pairs, respectively. By introducing the structural relational knowledge, Eq. (\ref{eq:rcd}) can effectively maximize the similarity between the outputs of the student and teacher.

To improve performance, we incorporate the naive logit distillation loss \(L_{\text{kd}} = \sum_{i=1}^{|\mc{D}|}\ell_\text{kl}\left(S_h(\bl{f}_i; \bl{w}_h), T_d(\bl{x}_i; \bl{w}_h)\right)\)~\cite{Hinton2015DistillingTK} with KL divergence \(\ell_\text{kl}\) and the classification loss \(L_{\text{cls}}=\sum_{i=1}^{|\mc{D}|}\ell_\text{ce}\left(S_h(\bl{f}_i; \bl{w}_h), y_i\right)\) with cross-entropy \(\ell_\text{ce}\), and defined the total discriminative loss \(L_{\text{dis}}\) as:
\begin{small}
\begin{equation}
  L_\text{dis}(\bl{w}_h;\mc{D}) = L_{\text{cls}} + 0.25 L_{\text{kd}} + 4.0 L_{\text{rcd}}.
\label{eq:disloss}
\end{equation}
\end{small}

\subsection{Module-Wise Training}

Due to the diversity between generative representations and discriminative representations, we train the student in a module-wise manner rather than the end-to-end training. First, the student backbone \(S_b\) is trained on massive low-resolution faces \(\{\hat{\bl{x}}_i\}_{i=1}^{|\mc{D}|}\) by minimizing the generative loss in Eq.~(\ref{eq:genloss}). This training is supervised by the pretrained generative model without face identities, thereby enabling the student to learn general and robust face representations. Then, \(S_b\) is fixed and the student head \(S_h\) is further trained on \(\mc{D}\) by minimizing the dicriminative loss in Eq.~(\ref{eq:disloss}). In this way, the complete structural knowledge is efficiently transferred from both generative and discriminative teachers, leading to a discriminative student \(S=\{S_b,S_h\}\). The training in the two distillation modules can be effectively performed with the back-propagation algorithm.

\section{Experiments}\label{sec:exp}
To validate the effectiveness of our \textbf{GDRD}, we train the student models on WebFace~\cite{yi2014learning}, evaluate on four benchmarks (LFW~\cite{LFWTech}, UCCS \cite{uccs}, TinyFace~\cite{cheng2018accv} and AR~\cite{ARFace1998}) and compare with the state-of-the-arts. To simulate the very low-resolution conditions and make comparison fair, we perform face detection and alignment, and resize the face images to \(112\times112\) for the discriminative teacher and \(16\times16\) for the student as well as generative teacher. We take the encoder of a pretrained diffusion model PGDiff~\cite{yang2023nips} as generative teacher and ArcFace~\cite{deng2019arcface} for discriminative teacher. We use the same network in \cite{zhang2024tcsvt} for  student, which includes convolutional layers of ResNet18 with channels of 128, 256, 512 along with ReLU fully-connected layers. We set the batch size to 96, the initial learning rate to 0.05 and the annealing rate to 0.1 to ensure the repeatability of the experiments. We fix the random seed at 7. Our models are implemented based on PyTorch with four Nvidia 3090 GPUs.

\subsection{Very Low-Resolution Face Verification on LFW}

\begin{small}
\begin{table}[tb]
\centering
\caption{Low-resolution face verification accuracy on LFW~\cite{LFWTech} under the resolution of 16×16. }
\label{tab:lfw}
\footnotesize  
\begin{tabular}{c|c|c}
\hline
Model              & Average Accuracy (\%) & Publication   \\ 
\hline
FaceNet~\cite{schroff2015facenet} & 90.25    & CVPR 2015  \\
CosFace~\cite{wang2018cvpr} & 93.80    & CVPR 2018  \\
ArcFace~\cite{deng2019arcface}  & 92.30    & CVPR 2019  \\  
MagFace~\cite{meng2021magface}  & 94.97    & CVPR 2021  \\  
\hline
SiameseFace~\cite{SiameseFace2023ICIIP}  & 92.41    & ICIIP 2023 \\
MobilenetV3-SE~\cite{MobilenetV3-SE2024ICCECE}&92.86&ICCECE 2024\\
\hline
SKD~\cite{ge2019low} & 85.87    & TIP  2019   \\  
HORKD~\cite{ge2020aaai}  & 90.03    & AAAI 2020  \\  
EKD~\cite{zhang2022tcsvt} & 91.71    & TCSVT 2022 \\  
RPCL~\cite{li2022deep}  & 94.98  & NN 2022    \\
WaveResNet~\cite{WaveResNet2023VCIP} & 93.94    & VCIP 2023  \\
CRD~\cite{CRD2023CEI} & 94.62    & CEI 2023   \\
ADSRAM~\cite{ADSRAM2023PRML}  & 93.08    & PRML 2023  \\
CCFace~\cite{CCFACE2023IJCB} & 92.87    & IJCB 2023  \\
CRRCD~\cite{zhang2024tcsvt}&95.25&TCSVT 2024 \\
\hline
\textbf{Our GDRD}   & \textbf{96.13} & -- \\
\hline
\end{tabular}
\end{table}
\end{small}
We conduct evaluation on LFW~\cite{LFWTech} and compare with 15 state-of-the-arts. We extract a 512$d$ feature embedding for each face image, calculate the similarity over 3000 positive pairs and 3000 negative pairs, check each pair with an optimal threshold, compute the correct predictions, and report the results in Tab.~\ref{tab:lfw}. We find that our GDRD delivers the best accuracy of 96.13\% and conclude some meaningful observations.  

First, our GDRD outperforms 4 state-of-the-art normal face recognizers (FaceNet, CosFace, ArcFace and MagFace). For example, ArcFace with ResNet50 reaches an accuracy of 99.82\% under standard resolution but significantly drops to 92.30\% under $16\times16$, which clearly highlights the importance of supplementing missing face knowledge for low-resolution recognition. 
Second, compared to typical distillation-based approaches (e.g., CRD, MobilenetV3-SE, ADSRAM, and WaveResNet), our GDRD gives a higher accuracy due to the distillation of both discriminative and generative features as well as the extraction of high-level relational contrastive knowledge. Thus, our approach surpasses sample-level knowledge distillation models such as SKD~\cite{ge2019low} and EKD~\cite{zhang2022tcsvt}, and also outperforms low-order relation knowledge models like HORKD~\cite{ge2020aaai} and multi-stream CNN models like SiameseFace. Third, our approach, which uses anchor-based high-order relational distillation, implicitly encodes margin-based discriminative representation learning, thereby outperforming RPCL~\cite{li2022deep}  that learns margin-based discriminative low-resolution face features. Consequently, the application of high-order relations in cross-resolution knowledge transfer not only enhances learning from the low-resolution domain but also improves efficiency in visual recognition tasks. 

\subsection{Very Low-Resolution Face Identification on UCCS}
We evaluate on UCCS~\cite{uccs} with the same setting to SKD~\cite{ge2019low}. UCCS (UnConstrained College Students) dataset is captured in real surveillance scenario covering various weather conditions and containing various occlusions. In experiments, we take a subset containing 180 subjects, 3918 training images and 907 testing images. We freeze the feature extraction part, adjust only the final softmax layer for 180 categories, finetune its parameters on training set, and then evaluate identification accuracy on testing set.  
\begin{small}
\begin{table}[tb]
\centering
\caption{Very low-resolution face identification accuracy on UCCS~\cite{uccs} under the resolution of 16$\times$16. }
\label{tab:uccs}
\footnotesize
\begin{tabular}{c|c|c}
\hline
Model               &Average Accuracy (\%)&          Publication   \\ \hline
CosFace~\cite{wang2018cvpr}&91.83& CVPR 2018\\
ArcFace~\cite{deng2019arcface}&88.73&CVPR 2019\\
MagFace~\cite{meng2021magface}&33.14&CVPR 2021  \\\hline
DirectCapsNet~\cite{DirectCapsNet2019ICCV}&95.81&ICCV 2019\\
IASR~\cite{IASR2020SPL}&89.73&SPL 2020  \\ \hline
SKD~\cite{ge2019low}&67.25&TIP 2019\\
HORKD~\cite{ge2020aaai}&92.11& AAAI 2020  \\  
EKD~\cite{zhang2022tcsvt} &93.85&TCSVT 2022  \\
RPCL~\cite{li2022deep}&95.13 &TCSVT 2022 \\  
CRRCD~\cite{zhang2024tcsvt}&97.27&TCSVT 2024 \\\hline
\textbf{Our GDRD}& \textbf{97.56}& \textbf{---} \\ \hline
\end{tabular}
\end{table}
\end{small}
As shown in Tab. ~\ref{tab:uccs}, our student achieves the best identification accuracy of 97.56\% on UCCS,  which surpasses the second CRRCD with an improvement of 0.29\%. Despite the lack of crucial recognition information, our student effectively enhances knowledge from generative model and extracts cross-resolution contrastive knowledge from discriminative model as well as high-resolution images, enabling the student to acquire discriminative representations.We also compared the identity clustering effects of discriminative features extracted during testing, as shown in Fig. \ref{fig:tsne}. It is evident that our model demonstrates significantly better clustering result. This indicates that our student is capable of mastering higher-order feature representations to enhance performance.
\begin{small}
\begin{figure}[ht]
\centering  
\begin{minipage}{0.24\textwidth}
        \includegraphics[width=\linewidth]{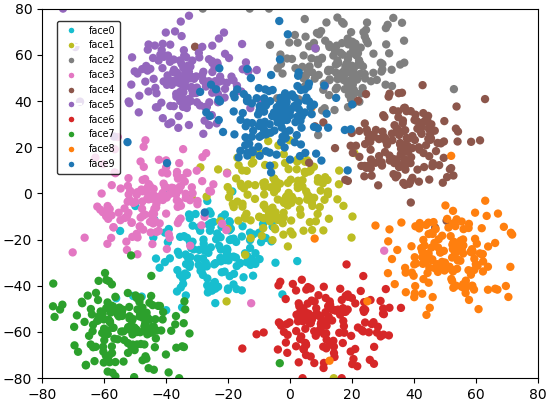} 
        \label{fig:leftimage}
    \end{minipage}
    \begin{minipage}{0.24\textwidth}
        \includegraphics[width=\linewidth]{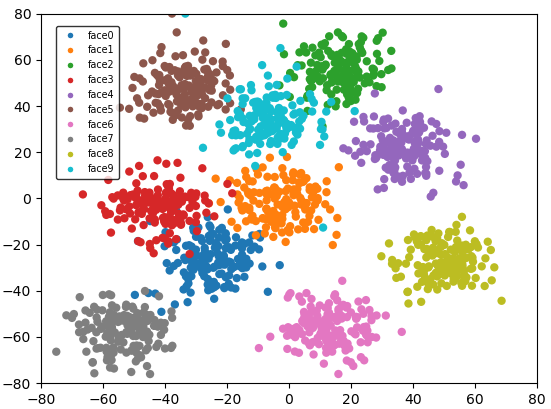} 
        \label{fig:rightimage}
    \end{minipage}
    \vspace{-4mm}
    \caption{t-SNE~\cite{van2008visualizing} visualization of representations extracted on UCCS by ArcFace (left) and our student (right).}
    \vspace{-3mm}
    \label{fig:tsne}
  \end{figure}
\end{small}
\subsection{Very Low-Resolution Face Retrieval on TinyFace}
In the experiments, we finetune our base model on TinyFace~\cite{cheng2018accv} training set and report the 1:N recognition performance on testing set in Tab.~\ref{tab:tinyface}. Our model achieves the highest retrieval results in Rank-1, Rank-5, and Rank-10, demonstrating strong recognition capabilities. Unlike other models, our approach implicitly learns clear boundaries between different classes under cross-resolution relations with the aid of a high-resolution teacher model. Our approach outperforms the super-resolution IASR model and the sample re-weighting SRW model. Compared to models like Mkmmd, HORKD, CCFace, CRD and CRRCD, our model demonstrates higher retrieval accuracy across all ranking metrics. These results highlight the effectiveness of our approach in learning discriminative and transferable representations, making it particularly suitable for low-resolution face recognition tasks.

\begin{small}
\begin{table}[tb]
\centering
\caption{Very low-resolution face retrieval performance on TinyFace~\cite{cheng2018accv} under the resolution of $16\times16$.}
\label{tab:tinyface}
\footnotesize
\begin{tabular}{c|c|c|c|c}  
\hline
Model               & Rank-1         & Rank-5     & Rank-10   & Publication \\ \hline
Baseline                & 31.21              & 47.25          & 48.21          & ACCV 2018 \\\hline
SKD~\cite{ge2019low} & 47.91          & 56.55          & 58.92          & TIP 2019    \\   
Mkmmd~\cite{mirzadeh2020improved} & 45.49     & 54.61          & 58.27          & AAAI 2020   \\
HORKD~\cite{ge2020aaai} & 45.49     & 54.80          & 58.26          & AAAI 2020   \\
SRW~\cite{ADSRAM2023PRML} & 49.25         & 58.49          & 61.23          & PRML 2023   \\
CRD~\cite{CRD2023CEI} & 49.00          & 57.12          & 60.73          & CEI 2023    \\ 
CCFace~\cite{CCFACE2023IJCB} & 48.03    &56.38    &59.17      & IJCB 2023  \\
CRRCD~\cite{zhang2024tcsvt} & 49.47    &58.98    &61.50      & TCSVT 2024  \\\hline
\textbf{Our GDRD} & \textbf{49.83}   & \textbf{59.14} & \textbf{61.97} & \textbf{---} \\ \hline
\end{tabular}
\end{table}
\end{small}
\subsection{Evaluation on Resolution and Occlusion Robustness}

\begin{small}
\begin{table}[tb]
\centering
\caption{Recognition accuracy (\%) on AR~\cite{ARFace1998} under three different scenarios and the resolution of $16\times16$.}
\label{tab:ar}
\footnotesize  
\begin{tabular}{c|c|c|c|c}
\hline
Model               &Illumination      & Eye Occ.   & Mouth Occ.&Publication   \\ \hline
ResNet34~\cite{he2016cvpr}     &75.32                            & 70.03  & 55.14&CVPR 2016   \\ 
MoblieNet~\cite{MobileNet2017}          &73.42                                  & 67.32 & 51.21&arXiv 2017  \\
IASR~\cite{IASR2020SPL}                 & 78.23                                 & 75.48  & 60.16&SPL 2020  \\\hline
SKD~\cite{ge2019low}                 &    73.70                  & 65.37  & 54.67&TIP 2019  \\  
EKD~\cite{zhang2022tcsvt}                  &  88.16                                & 80.72  &70.83&TCSVT 2022  \\ 
CCFace~\cite{CCFACE2023IJCB}  & 87.52    &81.34    &70.29&IJCB 2023      \\
\hline
\textbf{Our GDRD}       & \textbf{89.47}             & \textbf{84.33}  & \textbf{72.91}&\textbf{---} \\\hline
\end{tabular}
\end{table}
\end{small}
We first study the resolution robustness on UCCS. To this end, we trained three student models for low resolutions of \(8\times8\), \(16\times16\) and \(32\times32\), achieving the identification accuracy of 86.17\%, 97.56\% and 98.73\%, respectively. As expected, the resolution degradation reduces the recognition performance. However, our approach still can effectively recover the missing knowledge to achieve satisfying accuracy, \eg, comparable accuracy with ArcFace even under a lower resolution, implying its robustness.  

Beyond resolution, we further evaluate the occlusion robustness on AR dataset~\cite{ARFace1998} where the faces contain different expressions, illumination conditions and occlusions. We divide the dataset into three groups, based on illumination variations, eye occlusion, and mouth occlusion. As shown in Tab. \ref{tab:ar}, our approach achieves the best accuracy in all three scenarios. This demonstrates the superior robustness of our approach in handling illumination variations and partial occlusions. By employing a contrastive distillation approach, our approach effectively mitigates the effect of lighting and partial occlusions.

\subsection{Ablation Study}
\begin{small}
\begin{figure}[ht]
\centering  
\includegraphics[width=1.0\linewidth]{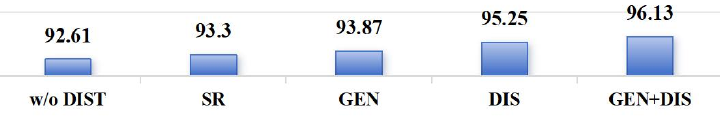} 
\caption{Ablation study of verification accuracy (\%) on LFW.}
\label{fig:ablation}
\end{figure}
\end{small}
After the promising performance is achieved, we study the effect of each module in our approach and report the results on LFW in Fig. \ref{fig:ablation}, where the students without distillation (w/o DIST), with super-resolving representations without distillation (SR), with only generative representation distillation (GEN), with only discriminative representation distillation (DIS) and with two distillation modules (GEN+DIS) deliver the accuracy of 92.61\%, 93.30\%, 93.87\%, 95.25\% and 96.13\%, respectively. The results indicate several meaningful observations: 1) the recovering of missing knowledge is difficult without distillation, 2) the generative representations is helpful in improving recognition, and 3) the generative representations need to be transformed into discriminative representations for further performance improvement, suggesting the effectiveness of our joint generative and discriminative representation distillation. 
\section{Conclusion}\label{sec:con}
In this paper, we propose a generative-discriminative representation distillation approach that successfully combines generative representation with cross-resolution knowledge distillation. This approach facilitates the transfer of higher-order relational knowledge between teachers and students, enhancing the transfer capabilities for very low-resolution face recognition. Extensive experiments on very low-resolution face recognition tasks have demonstrated the effectiveness and adaptability of our approach. Our future work will focus on integrating domain generalization and exploring the applicability of our approach to a broader range of visual understanding tasks.

\bibliographystyle{IEEEtran}
\bibliography{JunZheng-Zhang}

\begin{thebibliography}{10}
\providecommand{\url}[1]{#1}
\csname url@samestyle\endcsname
\providecommand{\newblock}{\relax}
\providecommand{\bibinfo}[2]{#2}
\providecommand{\BIBentrySTDinterwordspacing}{\spaceskip=0pt\relax}
\providecommand{\BIBentryALTinterwordstretchfactor}{4}
\providecommand{\BIBentryALTinterwordspacing}{\spaceskip=\fontdimen2\font plus
\BIBentryALTinterwordstretchfactor\fontdimen3\font minus \fontdimen4\font\relax}
\providecommand{\BIBforeignlanguage}[2]{{%
\expandafter\ifx\csname l@#1\endcsname\relax
\typeout{** WARNING: IEEEtran.bst: No hyphenation pattern has been}%
\typeout{** loaded for the language `#1'. Using the pattern for}%
\typeout{** the default language instead.}%
\else
\language=\csname l@#1\endcsname
\fi
#2}}
\providecommand{\BIBdecl}{\relax}
\BIBdecl

\bibitem{gong2019iccvw}
S.~Gong, Y.~Shi, and A.~Jain, ``Low quality video face recognition: Multi-mode aggregation recurrent network (marn),'' in \emph{Proc. IEEE/CVF Int. Conf. Comput. Vis. Worksh. (ICCVW)}, 2019, pp. 1027--1035.

\bibitem{chai2023cvpr}
J.~C.~L. Chai, T.-S. Ng, C.-Y. Low, J.~Park, and A.~B.~J. Teoh, ``Recognizability embedding enhancement for very low-resolution face recognition and quality estimation,'' in \emph{Proc. IEEE/CVF Conf. Comput. Vis. Pattern Recog. (CVPR)}, 2023, pp. 9957--9967.

\bibitem{schroff2015facenet}
F.~Schroff, D.~Kalenichenko, and J.~Philbin, ``Facenet: A unified embedding for face recognition and clustering,'' in \emph{Proc. IEEE Conf. Comput. Vis. Pattern Recog. (CVPR)}, 2015, pp. 815--823.

\bibitem{wang2018cvpr}
H.~Wang, Y.~Wang, Z.~Zhou, X.~Ji, D.~Gong, J.~Zhou, Z.~Li, and W.~Liu, ``{CosFace}: Large margin cosine loss for deep face recognition,'' in \emph{Proc. IEEE/CVF Conf. Comput. Vis. Pattern Recog. (CVPR)}, 2018, pp. 5265--5274.

\bibitem{deng2019arcface}
J.~Deng, J.~Guo, N.~Xue, and S.~Zafeiriou, ``Arcface: Additive angular margin loss for deep face recognition,'' in \emph{Proc. IEEE/CVF Conf. Comput. Vis. Pattern Recog. (CVPR)}, 2019, pp. 4685--4694.

\bibitem{kim2020groupface}
Y.~Kim, W.~Park, M.-C. Roh, and J.~Shin, ``{GroupFace}: Learning latent groups and constructing group-based representations for face recognition,'' in \emph{Proc. IEEE/CVF Conf. Comput. Vis. Pattern Recog. (CVPR)}, 2020, pp. 5620--5629.

\bibitem{ge2020efficient}
S.~Ge, S.~Zhao, C.~Li, Y.~Zhang, and J.~Li, ``Efficient low-resolution face recognition via bridge distillation,'' \emph{IEEE Trans. Image Process. (TIP)}, vol.~29, pp. 6898--6908, 2020.

\bibitem{biswas2011multidimensional}
S.~Biswas, K.~W. Bowyer, and P.~J. Flynn, ``Multidimensional scaling for matching low-resolution face images,'' \emph{Transactions on Pattern Analysis and Machine Intelligence(TPAMI)}, vol.~34, no.~10, pp. 2019--2030, 2012.

\bibitem{ge2019low}
S.~Ge, S.~Zhao, C.~Li, and J.~Li, ``Low-resolution face recognition in the wild via selective knowledge distillation,'' in \emph{IEEE Trans. Image Process. (TIP)}, vol.~28, no.~4, 2019, pp. 2051--2062.

\bibitem{zhu2019icassp}
M.~Zhu, K.~Han, C.~Zhang \emph{et~al.}, ``Low-resolution visual recognition via deep feature distillation,'' in \emph{IEEE Int. Conf. Acoustics, Speech and Sig. Process. (ICASSP)}, 2019, pp. 3762--3766.

\bibitem{yan2019iccvw}
M.~Yan, M.~Zhao, Z.~Xu, Q.~Zhang, G.~Wang, and Z.~Su, ``{VarGFaceNet}: An efficient variable group convolutional neural network for lightweight face recognition,'' in \emph{Proc. IEEE/CVF Int. Conf. Comput. Vis. Worksh. (ICCVW)}, 2019, pp. 2647--2654.

\bibitem{ge2019mm}
S.~Ge, S.~Zhao, X.~Gao, and J.~Li, ``Fewer-shots and lower-resolutions: Towards ultrafast face recognition in the wild,'' in \emph{ACM Int. Conf. Multimedia (MM)}, 2019, pp. 229--237.

\bibitem{ge2020aaai}
S.~Ge, K.~Zhang, H.~Liu, Y.~Hua, S.~Zhao, X.~Jin, and H.~Wen, ``Look one and more: Distilling hybrid order relational knowledge for cross-resolution image recognition,'' in \emph{Proc. AAAI Conf. Artif. Intell. (AAAI)}, 2020, pp. 10\,845--10\,852.

\bibitem{zhang2024tcsvt}
K.~Zhang, S.~Ge, R.~Shi, and D.~Zeng, ``Low-resolution object recognition with cross-resolution relational contrastive distillation,'' in \emph{IEEE Trans. Circuit Syst. Video Technol. (TCSVT)}, vol.~34, no.~4, 2024, pp. 2374--2384.

\bibitem{zha2019tcn}
J.~Zha and H.~Chao, ``Tcn: Transferable coupled network for cross-resolution face recognition,'' in \emph{IEEE Int. Conf. Acoustics, Speech and Sig. Process. (ICASSP)}, 2019, pp. 3302--3306.

\bibitem{rombach2022cvpr}
R.~Rombach, A.~Blattmann, D.~Lorenz, P.~Esser, and B.~Ommer, ``High-resolution image synthesis with latent diffusion models,'' in \emph{Proc. IEEE/CVF Conf. Comput. Vis. Pattern Recog. (CVPR)}, 2022, pp. 10\,684--10\,695.

\bibitem{yang2023nips}
P.~Yang, S.~Zhou, Q.~Tao, and C.~C. Loy, ``{PGDiff}: Guiding diffusion models for versatile face restoration via partial guidance,'' in \emph{Proc. Adv. Neural Inform. Process. Syst.}, 2023, pp. 32\,194--32\,214.

\bibitem{esser2024icml}
P.~Esser, S.~Kulal, A.~Blattmann \emph{et~al.}, ``Scaling rectified flow transformers for high-resolution image synthesis,'' in \emph{Proc. Int. Conf. Mach. Learn. (ICML)}, 2024, pp. 12\,606--12\,633.

\bibitem{yang2023iccv}
X.~Yang and X.~Wang, ``Diffusion model as representation learner,'' in \emph{Proc. IEEE/CVF Int. Conf. Comput. Vis. (ICCV)}, 2023, pp. 18\,892--18\,903.

\bibitem{ge2024icml}
S.~Ge, W.~Guo, C.~Li, J.~Zhang, Y.~Li, and D.~Zeng, ``Masked face recognition with generative-to-discriminative representations,'' in \emph{Proc. Int. Conf. Mach. Learn. (ICML)}, 2024, pp. 15\,242--15\,254.

\bibitem{Hinton2015DistillingTK}
\BIBentryALTinterwordspacing
G.~E. Hinton, O.~Vinyals, and J.~Dean, ``Distilling the knowledge in a neural network,'' in \emph{Proc. Adv. Neural Inform. Process. Syst. Worksh.}, 2015. [Online]. Available: \url{http://arxiv.org/abs/1503.02531}
\BIBentrySTDinterwordspacing

\bibitem{yi2014learning}
\BIBentryALTinterwordspacing
D.~Yi, Z.~Lei, S.~Liao, and S.~Z. Li, ``Learning face representation from scratch,'' \emph{arXiv preprint}, 2014. [Online]. Available: \url{http://arxiv.org/abs/1411.7923}
\BIBentrySTDinterwordspacing

\bibitem{LFWTech}
G.~B. Huang, M.~A. Mattar, T.~L. Berg, and E.~Learned-Miller, ``Labeled faces in the wild: A database forstudying face recognition in unconstrained environments,'' University of Massachusetts, Tech. Rep., 2008.

\bibitem{uccs}
M.~Günther, P.~Hu, C.~Herrmann \emph{et~al.}, ``Unconstrained face detection and open-set face recognition challenge,'' in \emph{International Joint Conference on Biometrics (IJCB)}, 2017, pp. 697--706.

\bibitem{cheng2018accv}
Z.~Cheng, X.~Zhu, and S.~Gong, ``Low-resolution face recognition,'' in \emph{Asian Conference on Computer Vision (ACCV)}, 2018, pp. 605--621.

\bibitem{ARFace1998}
A.~Martinez and R.~Benavente, ``The {AR} face database: Cvc technical report,'' Autonomous University of Barcelona, Tech. Rep., 1998.

\bibitem{meng2021magface}
Q.~Meng, S.~Zhao, Z.~Huang, and F.~Zhou, ``Magface: A universal representation for face recognition and quality assessment,'' in \emph{Proc. IEEE/CVF Conf. Comput. Vis. Pattern Recog. (CVPR)}, 2021, pp. 14\,220--14\,229.

\bibitem{SiameseFace2023ICIIP}
R.~Vachhani, S.~Mandal, and B.~Gohel, ``Low-resolution face recognition using multi-stream cnn in siamese framework,'' in \emph{Seventh International Conference on Image Information Processing (ICIIP)}, 2023, pp. 85--90.

\bibitem{MobilenetV3-SE2024ICCECE}
Y.~Li, Y.~Zhang, and W.~Chen, ``Occluded face recognition algorithm incorporating attention mechanism,'' in \emph{International Conference on Consumer Electronics and Computer Engineering (ICCECE)}, 2024, pp. 91--94.

\bibitem{zhang2022tcsvt}
K.~Zhang, C.~Zhang, S.~Li, D.~Zeng, and S.~Ge, ``Student network learning via evolutionary knowledge distillation,'' \emph{IEEE Trans. Circuit Syst. Video Technol. (TCSVT)}, vol.~32, no.~4, pp. 2251--2263, 2022.

\bibitem{li2022deep}
P.~Li, S.~Tu, and L.~Xu, ``Deep rival penalized competitive learning for low-resolution face recognition,'' \emph{Neural Networks (NN)}, vol. 148, pp. 183--193, 2022.

\bibitem{WaveResNet2023VCIP}
Y.~Lu and T.~Ebrahimi, ``Cross-resolution face recognition via identity-preserving network and knowledge distillation,'' in \emph{Visual Communications and Image Processing Conference (VCIP)}, 2023, pp. 1--5.

\bibitem{CRD2023CEI}
S.~Xu and S.~Tang, ``Low resolution face recognition based on contrast representa distillation,'' in \emph{Electronic Information Engineering and Intelligent Control Technology (CEI)}, 2023, pp. 274--278.

\bibitem{ADSRAM2023PRML}
S.~Tang and X.~Sun, ``Attention distillation with sample re-weighting and adaptive margin for low-resolution face recognition,'' in \emph{Pattern Recognition and Machine Learning (PRML)}, 2023, pp. 80--85.

\bibitem{CCFACE2023IJCB}
M.~S.~E. Saadabadi, S.~R. Malakshan, H.~Kashiani, and N.~M. Nasrabadi, ``Ccface: Classification consistency for low-resolution face recognition,'' in \emph{International Joint Conference on Biometrics (IJCB)}, 2023, pp. 1--10.

\bibitem{DirectCapsNet2019ICCV}
M.~Singh, S.~Nagpal, R.~Singh, and M.~Vatsa, ``Dual directed capsule network for very low resolution image recognition,'' in \emph{Proc. IEEE/CVF Int. Conf. Comput. Vis. (ICCV)}, 2019, pp. 340--349.

\bibitem{IASR2020SPL}
J.~Chen, J.~Chen, Z.~Wang, C.~Liang, and C.-W. Lin, ``Identity-aware face super-resolution for low-resolution face recognition,'' \emph{IEEE Signal Processing Letters (SPL)}, vol.~27, pp. 645--649, 2020.

\bibitem{van2008visualizing}
L.~van~der Maaten and G.~Hinton, ``Visualizing data using t-sne,'' \emph{J. Mach. Learn. Res. (JMLR)}, vol.~9, no.~86, pp. 2579--2605, 2008.

\bibitem{mirzadeh2020improved}
S.~Mirzadeh, M.~Farajtabar, A.~Li, N.~Levine, A.~Matsukawa, and H.~Ghasemzadeh, ``Improved knowledge distillation via teacher assistant,'' in \emph{Proc. AAAI Conf. Artif. Intell. (AAAI)}, 2020, pp. 5191--5198.

\bibitem{he2016cvpr}
K.~He, X.~Zhang, S.~Ren, and J.~Sun, ``Deep residual learning for image recognition,'' in \emph{Proc. IEEE Conf. Comput. Vis. Pattern Recog. (CVPR)}, 2016, pp. 770--778.

\bibitem{MobileNet2017}
\BIBentryALTinterwordspacing
A.~G. Howard, M.~Zhu, B.~Chen, D.~Kalenichenko, W.~Wang, T.~Weyand, M.~Andreetto, and H.~Adam, ``Mobilenets: Efficient convolutional neural networks for mobile vision applications,'' \emph{ArXiv}, 2017. [Online]. Available: \url{http://arxiv.org/abs/1704.04861}
\BIBentrySTDinterwordspacing

\end{thebibliography}
\end{document}